\documentclass[letterpaper, 10 pt, conference]{ieeeconf}  

\IEEEoverridecommandlockouts                              
\usepackage{graphics} 
\usepackage{epsfig} 
\usepackage{mathptmx} 
\usepackage{times} 
\usepackage{amsmath} 
\usepackage{amssymb}  
\usepackage{subfigure}
\usepackage{balance}
\usepackage{siunitx}
\usepackage{comment}
\usepackage{xcolor}
\usepackage{dirtytalk}
\usepackage{bm}
\usepackage{gensymb}
\usepackage[ruled, titlenotnumbered]{algorithm2e}
\usepackage{multirow}
\usepackage{cite}

\title{\LARGE \bf
Anisotropic Stiffness and Programmable Actuation for Soft Robots Enabled by an Inflated Rotational Joint
}
\author{Sicheng Wang$^{1}$, Eugenio Frias-Miranda$^{1}$, Antonio Alvarez Valdivia$^{1}$, and Laura H. Blumenschein$^{1}$
\thanks{$^{1}$Robust and Adaptive Design (RAAD) Lab, School of Mechanical Engineering, Purdue University, West Lafayette, IN 47906, USA}
\thanks{Email: \tt\small \{wang5239, efrias, alvar168, lhblumen\}@purdue.edu }
}

\begin{document}

\maketitle
\thispagestyle{empty}
\pagestyle{empty}

\begin{abstract}
Soft robots are known for their ability to perform tasks with great adaptability, enabled by their distributed, non-uniform stiffness and actuation. Bending is the most fundamental motion for soft robot design, but creating robust, and easy-to-fabricate soft bending joint with tunable properties remains an active problem of research.
In this work, we demonstrate an inflatable actuation module for soft robots with a defined bending plane enabled by forced partial wrinkling. This lowers the structural stiffness in the bending direction, with the final stiffness easily designed by the ratio of wrinkled and unwrinkled regions. 
We present models and experimental characterization showing the stiffness properties of the actuation module, as well as its ability to maintain the kinematic constraint over a large range of loading conditions. We demonstrate the potential for complex actuation in a soft continuum robot and for decoupling actuation force and efficiency from load capacity. The module provides a novel method for embedding intelligent actuation into soft pneumatic robots.
\end{abstract}
\section{Introduction}

Soft robots provide robust, yet simple solutions to many challenging problems in robotics, such as grasping objects adaptively~\cite{shintake18}, traversing highly confined environments~\cite{zhang19,hawkes17}, and interacting safely with human users~\cite{zhu22}. 
What enables these soft robotic solutions is how their compliance allows underactuated modalities of movement which can adapt the robot's motions based on expected or unexpected interaction. However, it is not typically beneficial for soft robot systems to have uniform actuation or uniform compliance in all directions, partially due to challenges in design, modeling, and control, and how out-of-plane compliance can reduce performance under external force~\cite{su22}. As such, one of the most common modalities of movement, shared between both soft and rigid robots, is the bending joint: bending is omnipresent in soft robots, found in grippers, soft arms, and crawlers, and the parallelism with rigid-bodied motion has allowed simplified modeling approaches~\cite{armanini23}.

Bending in soft robots is typically achieved by utilizing the interplay between the actuation energy and the structures or materials of the soft body, resulting in local strain differences and overall movement. The design of PneuNet actuators~\cite{mosadegh14} provides an illustrative example of this interaction: the inflatable elastomer chamber by itself provides uniform material extension under pressure, which the strain-limiting layer then directs~\cite{liu21}. More complex variations of the interaction, such as ones between multiple actuators, enable bi-directional bending~\cite{lin23}, spatial bending~\cite{webster10}, and sequential execution of actuation~\cite{olson22}. Strategic design of constraints and actuators expands the configuration space of the deformation, allowing programming the torsion along the body of a soft robot to form helices~\cite{wang18,starke17} and general curves~\cite{blumenschein22,wang24}. 

Since bending motions rely on production of strain difference, high strain materials, like elastomers, remain popular, especially in fluidic systems. From a design perspective, complex programming on motion becomes possible from material alone using multi-material composites~\cite{nguyen20} and local property tuning~\cite{nakajima20,chen23}.
However, the high compliance of these materials makes rejecting external forces difficult, especially in out-of-plane directions that are not actuated~\cite{su22,good24}. As well, without specialized mitigation, joining materials whose Young's modulus differ across magnitudes introduces stress concentration or causes delamination at the interface~\cite{arase24}. The non-linear mechanics of the elastomer material create barriers for accurately predicting the deformation, which is typically solved with finite-element methods or learning-based simulation~\cite{qin24}.

\begin{figure}[t]
    \centering
    \includegraphics[width=.79\columnwidth]{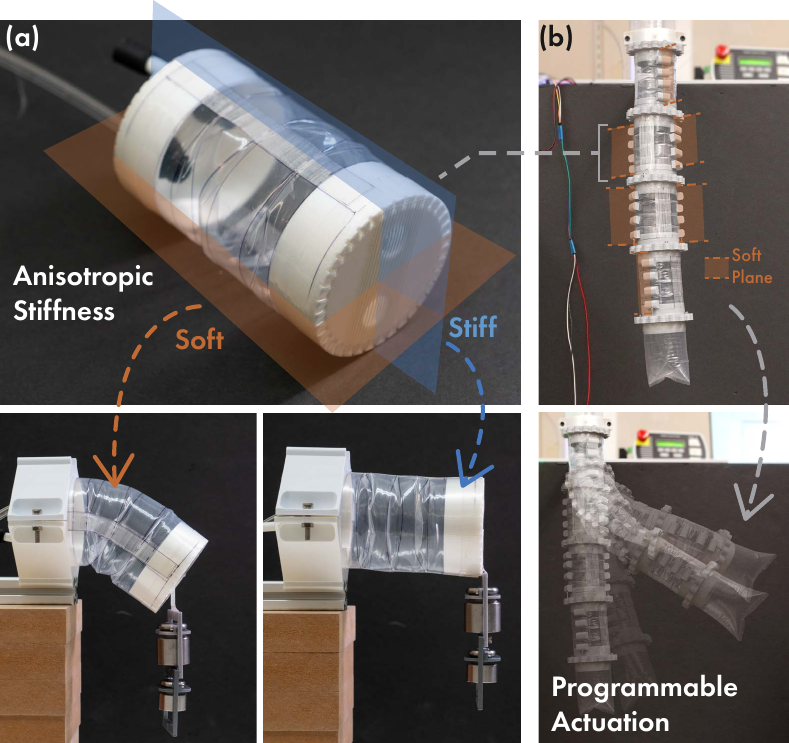}
    \caption{(a) The proposed inflated rotational joint features a bending plane with low stiffness. Bending motion will occur within the soft plane only unless the load causes buckling along the stiff plane. (b) An example application of the multiple joints, in which the relative stiffness and bending constraint of the joints enables defined kinematics and motion sequence.}
    \label{fig:firstFig}
    \vspace{-2em}
\end{figure}

Soft robot designs based on thin-filmed inflatable structures, which can be fabricated at comparable scale and operating pressure with their elastomer-based counterparts, offer an alternative mechanism for creating the structures to induce bending. Since the thin materials can only sustain tension, not significant compression, bending occurs on inflated beams by buckling when the local axial tension is lost~\cite{leonard60}, which may occur as a result of actuation~\cite{blumenschein22}, enforced pinch points~\cite{usevitch20, jiang24} wrinkling regions~\cite{voisembert13, hwee23}, or fully deflated "hinges"~\cite{kim24}. 
Many inflatable robot designs feature well-defined jointsms which have kinematics similar to rigid-bodied robots, defined by joints, links, and actuators embedded at the joints~\cite{voisembert13,kim24}. This architecture is informed by the snap-through transition between the stiff and compliant state at buckling, which makes the overall response function more piecewise than continuous. As a result, stiffness tuning of an inflated hinge is typically achieved via active control of the input pressure or antagonism with actuators~\cite{abrar21} within the stiff or buckled regime, and remains challenging from a design perspective.

In this paper, we present an approach to shape the bending behavior of a soft robot, by selectively imposing a zero-axial-stress (wrinkled) condition along a thin-filmed inflatable beam to create anisotropic bending stiffness (Figure~\ref{fig:firstFig}). While based on thin-filmed material, the approach addresses shortcomings in both elastomer and thin-film-based current solutions. The proposed mechanism has the ability to maintain the kinematic constraint of in-plane bending over a large range of loads when sufficiently pressurized, and hence comes the concept of \textit{Inflated Rotational Joint}. The proposed approach opens a design space for the stiffness difference between the bending directions, which, in combination with appropriate actuator routing, enables programmability in both kinematics and motion sequence for multiple units. The proposed inflated rotational joint structure provides a simple and effective way to shape the stiffness of soft robots without requiring components of different stiffness moduli.

The remainder of this paper is structured as follows. First, we develop the concept of the inflatable rotational joint and provide an example of its implementation with responses to external loads. We then examine the behavior of the joint as part of an autonomous system: beginning with a single unit with internally routed tendon, followed by a system with multiple of these units connected in series. Finally, we demonstrate an underactuated gripperthe rotational joint, showing its capability to realize the movement for grasping while resisting out-of-plane loads.

\section{The Inflated Rotational Joint}
\subsection{Concept and Realization}
\label{sec:anisoConcept}

\begin{figure}[tb]
    \centering
    \includegraphics[width=.9\columnwidth]{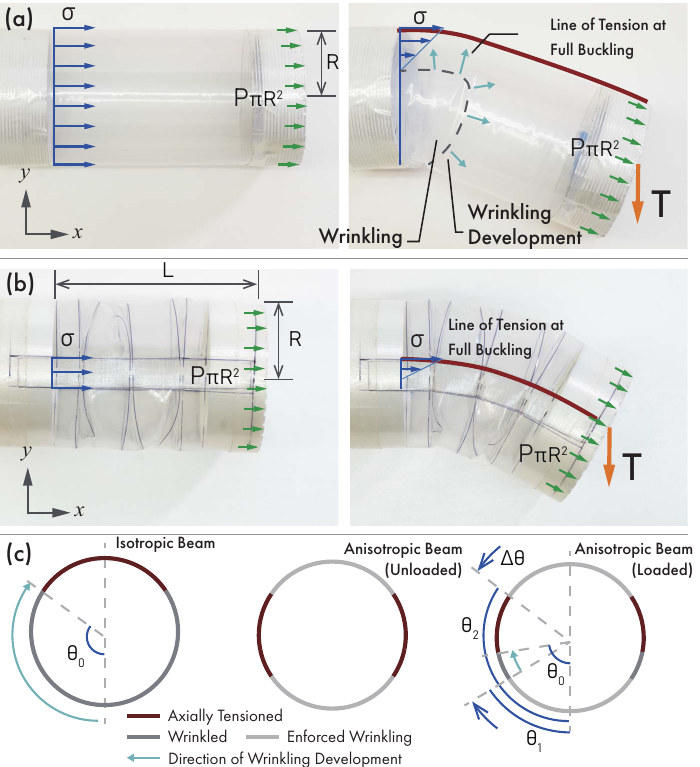}
    \caption{The load and axial tension in (a) isotropic inflated beam and (b) anisotropic inflated beam. (c) Definition o the tensioned and wrinkled regions' locations shown in a section view on the $z-y$ plane.}
    \label{fig:concepts}
    \vspace{-2em}
\end{figure}

To design our \textit{Inflated Rotational Joint}, we begin by reviewing the mechanics of an inflated beam made of flexible, thin-film material~\cite{comer63,leonard60}. At any given section plane, the force exerted by the pressure at the tip of the beam must be balanced by the tension in the thin-film membrane. This tension starts out uniform around the section when the beam is undisturbed. We consider an inflated beam with the proximal end fixed, and a load applied in the $-y$ direction at $x=L$ (Figure~\ref{fig:concepts}(a)). This applied load induces a linear axial stress distribution at each section: the tensile stress is reduced at the bottom and increased at the top to balance both the pressure load and the new moment due to the applied load. As the load is increased, the tension in the bottom region of the section drops further until it reaches zero tension, allowing wrinkles to develop in the material and bending in the structure. This wrinkling condition will first develop at the section experiencing the highest moment, which will be the proximal end for a uniform beam. As the applied load is increased, the wrinkled region expands upward around the circumference. Finally, for a maximally bent inflated beam, considered fully buckled, the only region of the section that remains under tension is a narrow point at the top, which will balance all of the load from the tip pressure, $\pi P R^2$. A beam in this tension configuration exhibits the maximal restoring moment, as given by~\cite{leonard60}:
\begin{equation}
    M_{max} = \pi P R^3,
    \label{eqn:beamMoment}
\end{equation}
where $P$ and $R$ are the pressure and beam radius.

For the isotropic beam, we can understand the maximum moment (Equation (\ref{eqn:beamMoment})) as resulting from the axial load due to pressure being balanced by the single line of tension acting at a lever arm of $R$ with respect to the bending plane. If we instead change the structure to move that final point in tension to be closer to the central axis, the maximum restoring moment will decrease. 
Enforcing this lower maximum height for the region axially in tension implies that the membrane further up cannot support an axial load. We achieve this by manually creating wrinkles in this section of the beam so that it can extend (Figure~\ref{fig:concepts}(b)). Doing this symmetrically ensures the beam remains undeformed at no load. 
While shifting the tensioned location decreases the restoring moment in the bending plane passing through enforced wrinkling, the restoring moment in the orthogonal bending plane, which passes through both axially tensioned region, is unchanged. Thus, the beam has different bending stiffness along the two principal bending planes, hereafter referred to as the \textit{soft} and \textit{stiff} plane respectively.
We manufactured the joints in this study from flattened thin-filmed LDPE tubing (ULine, IL, USA) of various widths and thicknesses. The manufacturing process is shown in Figure~\ref{fig:modelingResults}(a).

\subsection{Buckling Behavior Modeling}
\label{subsec:buckling}

\subsubsection{Stiffness Estimation from Lateral Load}

For a more rigorous derivation of the anisotropic stiffness, we focus on an analytical model for the joint. A similar analysis using finite-element modeling was shown in Voisembert et al.~\cite{voisembert13} Comer and Levy's inflated beam model~\cite{comer63}, the stress, $\sigma$, in a cantilevered inflated beam will satisfy
\begin{equation}
        Tx=-2\int^{\theta_2}_{\theta_1}t \sigma R^2 \cos\theta d\theta \\
        \label{eqn:forceIntegral1}
\end{equation}
\begin{equation}
    P\pi R^2 = 2\int^{\theta_2}_{\theta_1} t\sigma R d\theta,
    \label{eqn:forceIntegral2}
\end{equation}
where $T$ and $x$ are the magnitude and location of the applied load respectively, $t$ is the thickness of the film. The variable $\theta$ represents a location on the circular section profile (Figure~\ref{fig:concepts}(c)), in this case integrated between $\theta_1$ and $\theta_2$, which delineates the limits of the axially tensioned interval. Note we modify the upper and lower bound of the integral to be an unspecified combination of [$\theta_1$,$\theta_2$], as opposed to [$0$,$\pi$] in the original equations~\cite{comer63}, to reflect the partially-wrinkled condition. The tensile stress distribution, $\sigma$, has a linear profile, which relates to $\theta$ by:
\begin{equation}
    \sigma = \frac{\cos\theta_0-\cos\theta}{1+\cos\theta_0}\sigma_M,\;\;\mathrm{for} \;\;\theta_2 > \theta_0 \geq \theta_1
\end{equation}
where $\sigma_M$ is the maximum tensile stress, and $\theta_0$ is the location of the upper boundary of the lower wrinkled region, as shown in Figure~\ref{fig:concepts}(c), which has advanced farther up from the enforced wrinkling region due to the applied load. Integrating Equations~(\ref{eqn:forceIntegral1})-(\ref{eqn:forceIntegral2}), and solving for $\sigma_M$, we obtain the moment due to the tip load, $M=Tx$:
\begin{equation}
\begin{split}
    M &= P\pi R^3 f(\theta_0) \\
    &= P\pi R^3\frac{\sin2\theta_0+\sin2\theta_2+2(\theta_2-\theta_0)-4\cos\theta_0\sin\theta_2}{4[(\theta_2-\theta_0)\cos\theta_0-\sin\theta_2+\sin\theta_0]}.
    \label{eqn:fullMomentEqn}
\end{split}
\end{equation}
We can notice that this maximum moment is a scaled version of Equation~(\ref{eqn:beamMoment}) with a scale factor that is a function of $\theta_0$. It can be shown that $\lim_{\theta_0\rightarrow\theta_2}f=-\cos(\theta_2)$. If the regions of enforced wrinkling have equal width, we can alternatively express the maximum buckling moment in terms of the width of the tensioned region, $\Delta\theta=\theta_2-\theta_1$:
\begin{equation}
    M_{max} = \pi P R^3 \sin({\Delta\theta}/{2}),
    \label{eqn:beamMomentPartial}
\end{equation}
which shows that the restoring moment of the beam with forced wrinkling scales with the location of the highest point capable of sustaining tension, confirming Section~\ref{sec:anisoConcept}. Further, Equation~(\ref{eqn:beamMomentPartial}) indicates that the width of the axially tensioned region is the primary design parameter for tuning the relative stiffness of the bending planes.

\begin{figure}[tb]
    \centering
    \includegraphics[width=.94\columnwidth]{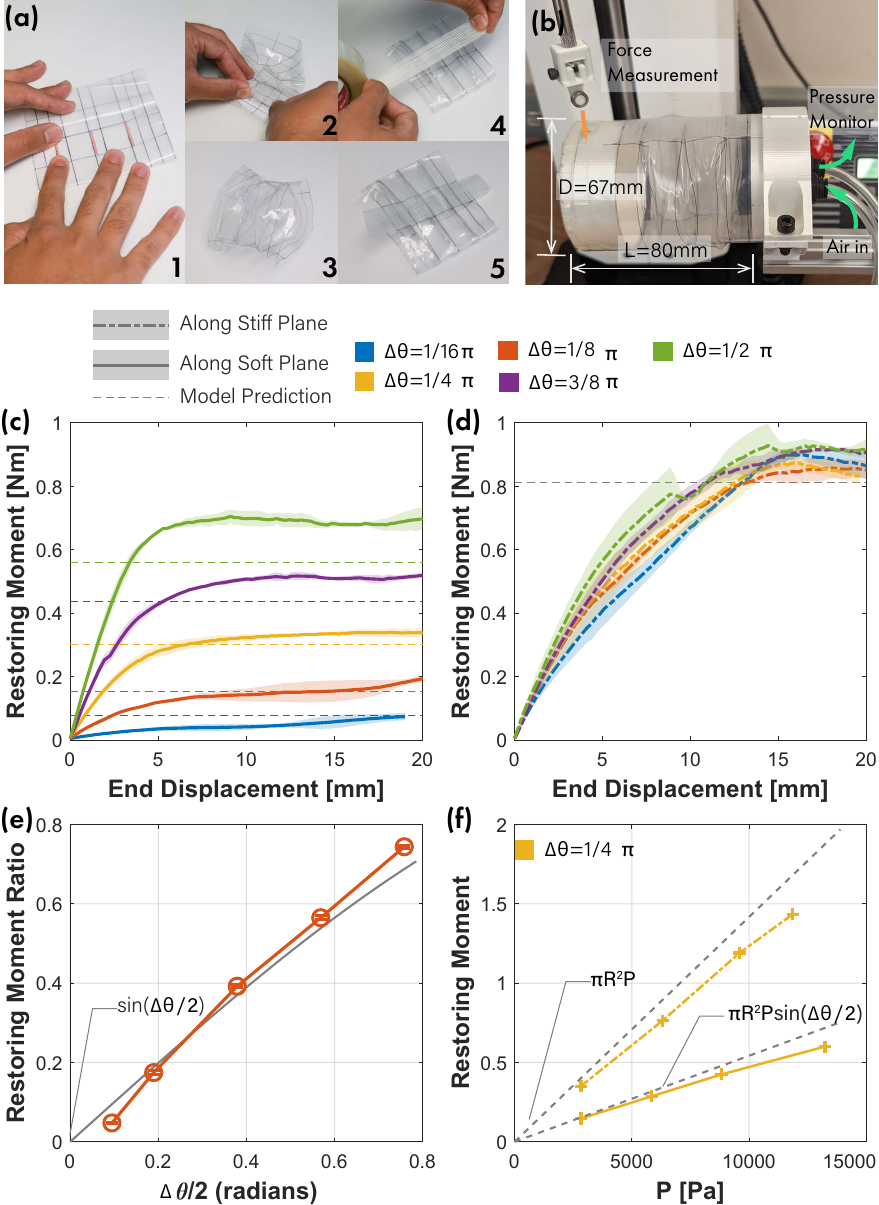}
    \caption{(a) The partially wrinkled condition can be enforced by shortening a piece of thin-filmed tubing by folding and using fiber-reinforced tape as a length constraint. (b) Testing setup for restoring moment verification. (c),(d) Restoring moment at various tip displacements, with the shaded region representing one standard deviation uncertainty. (e) Ratio of the restoring moment values along the planes for each axially tensioned region width, $\Delta\theta$. (f) Stiffness values along the stiff and soft bending planes for $\Delta\theta=\pi/4$ as a function of pressure.}
    \label{fig:modelingResults}
    \vspace{-1.8em}
\end{figure}

\subsection{Restoring Moment Model Verification}
\label{sec:momentExp}
Now that we have established the fundamental design and model of the anisotropic beam, we seek to validate its behavior through experimental characterization of the restoring moment as the width of the axially tensioned region varies. We fabricated inflated rotational joint samples with LDPE thin-film tubings (50~$\mu$m thickness, 67~mm inflated diameter) and rigid ends fabricated with a FDM printing process (PLA, 100\% infill), each with symmetric axially tensioned regions of width 6.35~mm, 12.7~mm, 25.4~mm, 38.1~mm, and 50.8~mm. This corresponds to a $\Delta\theta$ ranging from $\pi/16$ to $\pi/2$. The tensioned regions are enforced with a composite of adhesive tape (TrueTape LLC., CO, USA) and 50~$\mu$m Dynemma® fiber-reinforced film (Ripstop-by-the-Roll, NC, USA). All samples are 60~mm in length, with the wrinkled regions having a 33.3\% surface strain. The samples are installed cantilevered on a fixture, with the free end subjected to vertical loading from a force-displacement measurement device (EM-303, MR03-5, and M5I, Mark-10 Inc, NY, USA) at 80mm from the clamping. The load-displacement relation is measured along both the soft and stiff planes while the test sample is inflated to a pressure of 6.89~kPa with an electronic pressure controller (QB3, Proportion-Air, IN, USA).  Figure~\ref{fig:modelingResults}(b) shows the setup. 

Figure~\ref{fig:modelingResults}(c),(d) shows the load-displacement obtained in the experiment, averaged from the results for each orientation and tensioned region width in three repetitions. The result shows the restoring moment generally confirms the prediction of the model. The soft-plane results (Figure~\ref{fig:modelingResults}(c)) indicates greater accuracy in predicting the restoring moment at small $\Delta\theta$, with increasing underprediction of the stiffness at higher $\Delta\theta$. This suggests potential effects of increased stiffness in manufacturing the joints by taping. The restoring moment at the stiff region (Figure~\ref{fig:modelingResults}(d)) shows little effect of the tensioned region width, as is expected in the model. Figure~\ref{fig:modelingResults}(e),(f) demonstrates the scaling relationship of the stiffness with axially-tensioned region width ($\Delta\theta$) and pressure ($P$) respectively. Figure~\ref{fig:modelingResults}(e) gives a non-dimensionalized comparison of the restoring moment across the two bending planes. For each tensioned region width, we computed the ratio of restoring moment associated with the planes at buckling, as the average value of the minimal-slope interval of the moment-displacement curve. Figure~\ref{fig:modelingResults}(f) shows a bending test performed with a $P$ between 6.89 to 27.6~kPa, the maximum that the samples can handle before failure.
The result confirms the prediction from Equation~(\ref{eqn:beamMoment}) and (\ref{eqn:beamMomentPartial}), which suggests $\sin\Delta\theta/2$ gives the stiffness ratio between the two bending planes. With the result, if each of the regions in tension is reduced to a single line ($\Delta\theta\to0$), the module would have near-zero stiffness on the soft bending plane, and the neutral state becomes unstable due to the tip pressure load. With a large $\Delta\theta$ or low pressure, the stiffness difference between the bending planes would be too small to enforce an effective bending kinematics constraint.


\section{Actuated Joint Characterization}

\subsection{Actuation by Internal Tendon}

\begin{figure}[tb]
    \centering
    \includegraphics[width=.8\columnwidth]{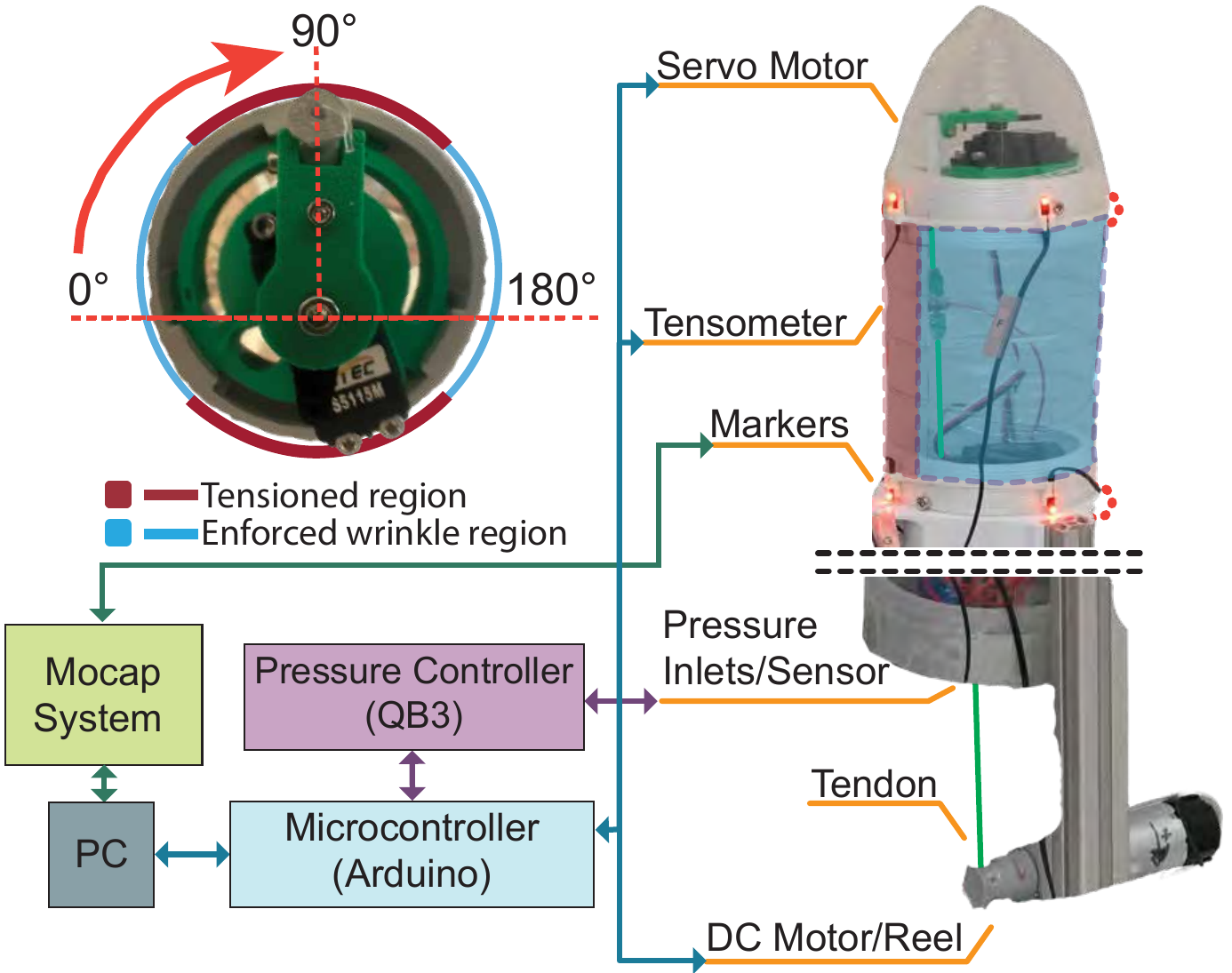}
    \caption{Experimental setup for tendon actuation. Motors control position and tension of the tendon, which is measured by a tension transducer. Eight motion capture markers track position and pressure is kept constant.
}
    \label{fig:singleUnitSetup}
    \vspace{-1.5em}
\end{figure}

To investigate how anisotropic stiffness of the \textit{Inflated Rotational Joint} affects actuation, we systematically apply tendon actuation to the joint. A short length of anisotropic stiffness beam with a single internal actuation tendon is built. Two servo motors (FS5115M Servo, FeeTech, Guangdong, China) at the top and bottom rotate between $-90^{\circ}$ and $90^{\circ}$, altering the tendon’s routing and adjusting the actuation lever arm. A DC motor (Pololu 25D Metal Gearmotor, NV, USA) controls the tendon length to actuate the module. Structural components within the joint were 3D printed (PETG, Bambu Labs). Pressure was regulated at 6.9~kPa by an electronic pressure controller (QB3, Proportion-Air, Inc., IN, USA), and a tension transducer (LCM100, FUTEK, CA, USA) measured tendon tension (Figure~\ref{fig:singleUnitSetup}).

\subsection{Experiment Design}

The experiment measured the restoring moment produced by the joint as the angular configuration of the tendons varied. Servo motors were adjusted in $20^\circ$ increments from $0^\circ$ to $180^\circ$. For each configuration, data collection included inputs from the encoder, pressure sensor, tension sensor, and motion capture system. A motion capture system (Impulse X2E System, PhaseSpace, San Leandro, CA) tracked eight markers on the joint, allowing calculation of the plane’s top and bottom center point, this determined buckling direction. An experiment with an isotropic joint was performed for comparison to assess the impact of anisotropy.


\subsection{Response to Actuation Input}

\begin{figure}[h]
    \centering
    \includegraphics[width=.85\columnwidth]{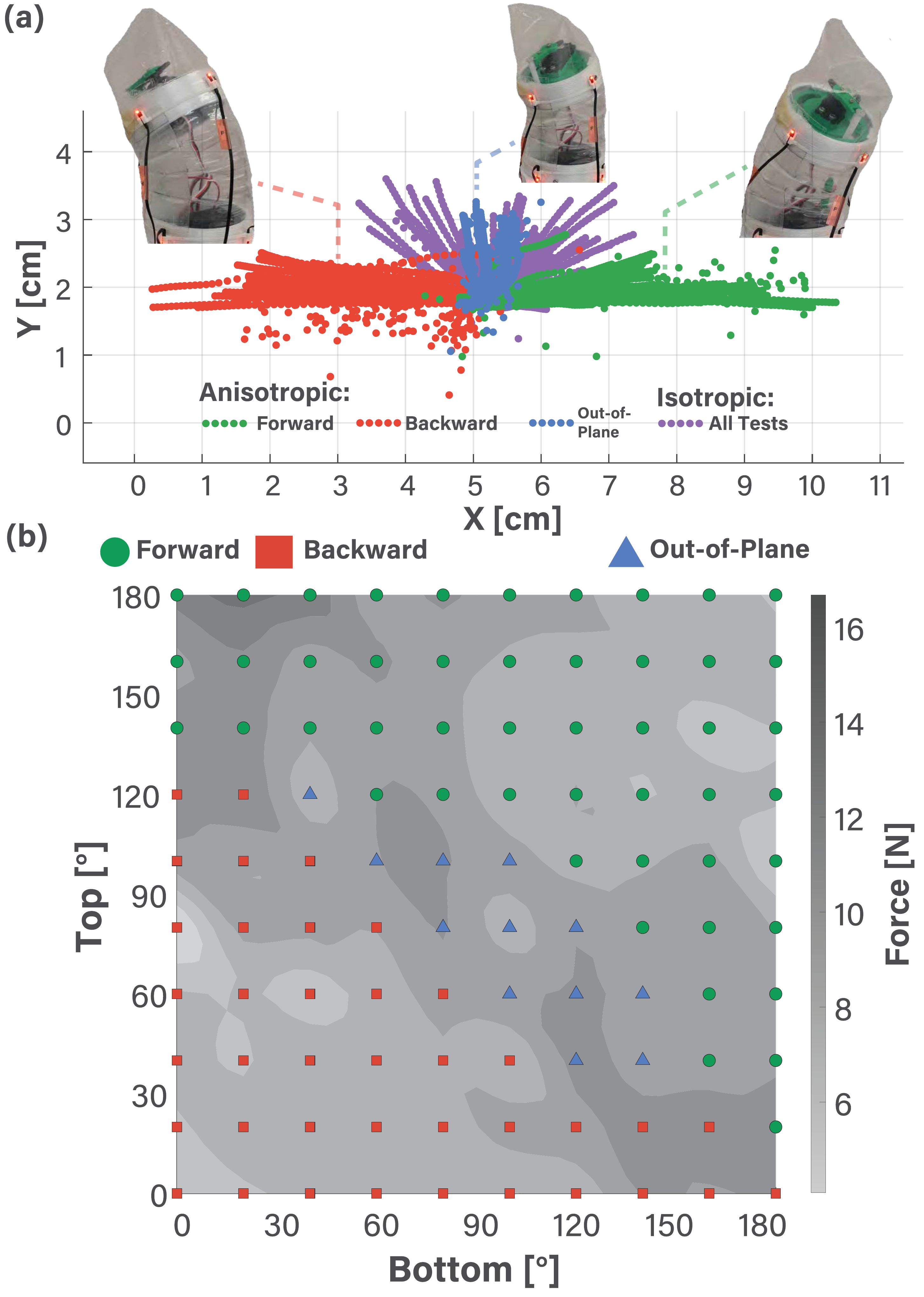}
    \caption{Resulting behavior from tendon actuation of anisotropic and isotropic units. Behaviors are categorized by unit type and primary bending direction. (a) Position of tip throughout the experiment in the XY plane, showing highly clustered bending in the soft plane for the anisotropic joint, compared to widely varied direction in the isotropic joint. (b) Tendon tension as a function of servo motor angle from $0^{\circ}$ to $180^{\circ}$ and bending behavior.}

    \label{fig:singleUnitResults}
    \vspace{-2em}
\end{figure}

The experiment results, illustrated in Figure~\ref{fig:singleUnitResults}(b), reveal a relationship between the angular configurations of the top and bottom servo motors and the buckling behavior of the \textit{Inflated Rotational Joint}. The anisotropic (wrinkled) joint exhibited a lower average buckling force of 10.45~N, with a maximum force of 23.23~N, indicating that the joint can buckle more easily, requiring less force when the tendons are aligned parallel to the central axis and close to the soft plane. In contrast, the isotropic (non-wrinkled) joint displayed a higher average buckling force of 16.47~N, with a maximum force of 31.72~N, showing its greater stiffness and resistance to bending.

These findings suggest that by introducing wrinkles into the joint's design, the anisotropic configuration reduces the forces needed for deformation and maintains bending in the desired plane over a large range of off-axis tendon loads (Figure~\ref{fig:singleUnitResults}(a)). Additionally, the tension to buckle varies as the tendon routing is adjusted, to the routing path can specify both the force required to induce buckling and the bending direction of an anisotropic joint. This is due to the tendon alignment influencing the moment arm of the applied tensile load, tailoring the beams response to the tension. This can offer greater design flexibility when designing underactuated systems that rely on tendon-driven actuation.

\section{Multi-Unit Demonstration}
The \textit{Inflated Rotational Joint} allows the bending direction and moment to be designed based on the structure and tendon path. By connecting multiple inflatable joints in series, we can create a reconfigurable robotic system capable of both predictable bending and actuation sequencing. This section discusses the development of a multi-unit system and how it facilitates the programming (and re-programming) of bending motion and actuation sequencing. 

\subsection{Design of Multi-Unit Prototype}

\begin{figure}[tb]
    \centering
    \includegraphics[width=.77\columnwidth]{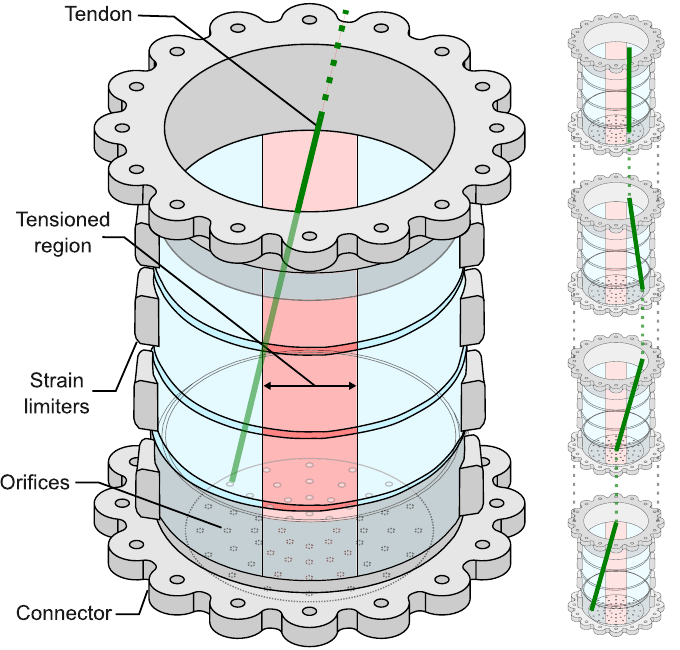}
    \caption{Illustration of the reconfigurable and modular inflatable joint design. The unit features the tensioned region, strain limiters, connectors, and orifices for tendon routing. Illustrations on the right show examples of tendon routing for multiple units connected in series, demonstrating the system's modularity and reconfigurability.}
    \label{fig:multiUnitDesign}
    \vspace{-1.5em}
\end{figure}

The multi-unit system is designed as a reconfigurable, modular body. The units are inflatable joints connected in series, forming an adaptable structure. These units follow the concept described in Section~\ref{sec:anisoConcept}, with the addition of two 3D-printed connectors (PETG, Bambu Labs), one at each end, as shown in Figure~\ref{fig:multiUnitDesign}. The connectors are designed to interface with one another while creating an air-tight connection between units with an o-ring. One plate of each pair includes orifices to define the tendon routing. M3 fasteners are used to secure the units in series. 

The connectors enable reconfigurability and modularity by allowing quick connection, disconnection, or rotation of the units to modify their arrangement. The tendon routing orifices allow for modification of the tendon's lever arm and the resulting bending moment from tendon tension, controlling the direction of bending and tension required to bend. However, the location of the tendon is not the only factor in determining the bending behavior. Per Equation~\ref{eqn:beamMomentPartial}, the width of the taped region tunes the stiffness of the units. For our modular units, we use three different taped region widths: 12.7, 25.4, and 38.1~mm. 
Since the modular system allows for modifying these factors independently across multiple units, 
we extend the models to predict , and hence design, the order in which they bend, creating actuation sequencing. 

\subsection{Actuation Sequencing}

\begin{figure*}[tb]
    \centering
    \includegraphics[width=0.95\textwidth]{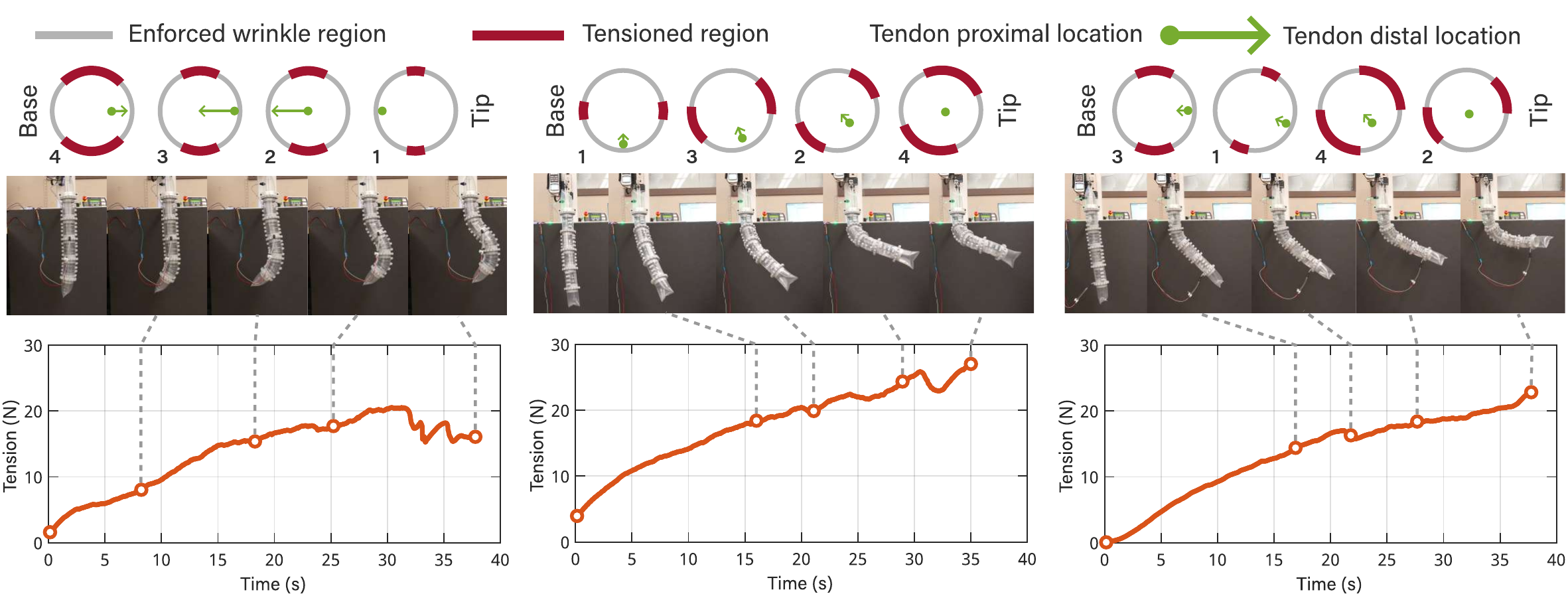}
    \caption{Actuation sequencing of the multi-unit system. The diagrams on the top row show the enforced wrinkle regions, tensioned regions, and tendon routing configurations for each unit in the series, with the order of actuation indicated by the numbers. The middle row presents images of the resulting sequences. The bottom row shows tension over time for each demonstration, marking the points shown in the actuation sequence, highlighting the correlation between the design variables and the resulting sequence of bending motion across units as the tension increases.}
    \label{fig:multiUnitResults}
    \vspace{-1.5em}
\end{figure*}

The sequence in which each joint bends is important for coordinating the actuation of multiple units with minimal inputs. This sequence is a combination of tendon routing, relative rotation, and the stiffness of each unit. Figure~\ref{fig:multiUnitResults} shows how four units can be reconfigured to create multiple different behaviors. The ordering of the units with different-sized tensioned regions, the orientation of each unit, and the routing of tendons from top to bottom are changed to achieve specific configurations. 

In the first demonstration (Figure~\ref{fig:multiUnitResults} left), we show motion in a 2D plane the tensioned regions of the units. Here, the bending sequence begins with the bottom unit, which has the smallest tension constraint and largest lever arm. The second and third demonstrations (center and right) show the same final3D helical shape, but with different bending sequences. In one configuration, the units are arranged in increasing sizes of tension constraints from top to bottom, so the motion starts from the base and ends at the tip. In the rightmost demonstration, the order of the units results in the second unit actuating first. Tendon placement in each configuration was determined heuristically since the constraints imposed by the width of the tensioned region affect the actuation sequence more: units with smaller tensioned regions require less tension to bend, and therefore, they are actuated first in the sequence. 
It is important to note that in all the configurations shown, two units have the same tension constraints (25.4~mm). The tendon position and resulting lever arm were not always sufficient to predict which of these segments would actuate first. We believe further modeling of the moment generated by non-uniform tendon routing  needed to refine our understanding of howtendon routing sequences actuation. 


\section{Application to Gripper Design}
\begin{figure}[h!]
    \centering
    \includegraphics[width=.85\columnwidth]{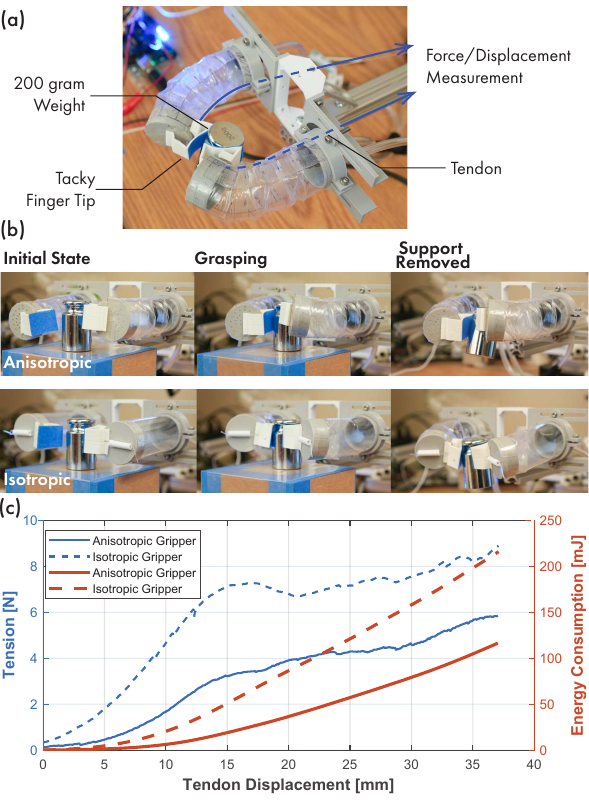}
    \caption{(a) The setup for a grasping actuation force and energy measurement. (b) Grasping motion sequence of the experiment, showing the anisotropic gripper (top) and isotropic design (bottom). (c) The tension in the tendon and the cumulative energy consumption during a grasp.}
    \label{fig:finger}
    \vspace{-2.5em}
\end{figure}

Finally, we demonstrate a soft gripper design from our bending units to highlight the benefits of the anisotropic stiffness. Targeting adaptability and stability in grasping, low in-plane stiffness helps with passively adapting to objects with few active degrees of freedom and lowering the energy required to actuate the grasp, while high out-of-plane stiffness increases load capacity and grasp stability. Therefore, we should expect that anisotropic stiffness will result in the same or better grasp performance.

We made a two-finger gripper with single bending units in two ways: one with taped wrinkles (12.7~mm width tape) to create anisotropic stiffness and the other without modification to the tube structure. Both sets of fingers are fabricated from 25~$\mu$m thickness, 35~mm diameter, and 90~mm length LDPE thin-filmed tube. Each finger features a 3D-printed flexible finger tip (NinjaFlex TPU, 20\% infill) with a tacky surface (Scotch Tape, 3M). A tendon made of unbraided nylon wire is routed on the surface of each finger and is actuated and measured by a force-displacement measurement device (Mark-10, NY, USA). In the demonstration, a 200~g object is first positioned within the gripper, the gripper then actuates to grasp the object and the support is removed while the object remains in the gripper (Figure~\ref{fig:finger}(b)). 

While both grippers succeed with an internal pressure of 6.89~kPa, the isotropic gripper requires more tendon tension, expends more energy, and still deflects more in the out-of-plane direction (Figure~\ref{fig:finger}(b)-(c)). Overall, the lower stiffness in the bending plane in the anisotropic design results in a 46\% lower energy cost. The higher out-of-plane deformation in the isotropic beam is likely because the wrinkles produced around the circumference by bending reach that bending plane, effectively lowering stiffness slightly. With earlier results (Figure~\ref{fig:modelingResults}(f)), we expect these benefits over a large range of pressure inputs and loads.

\section{Conclusion}
In this paper, we presented the design and analysis of an \textit{Inflated Rotational Joint}. The model and experiment results showed anisotropic stiffness with a tunable ratio of in- and out-of- plane stiffness. The joint can retain its kinematically constrained bending over a large range of loading conditions. With multiple units, this property creates actuation sequencing and more efficient force application. 

The proposed joint has an intuitive design space with predictable performance of design elements. Future work will refine the modeling to better understand the interaction between serially connected units, and will exploit the anisotropic stiffness for robot designs that robustly adapt to external interaction.


\balance
\vspace{-0.6em}
\bibliographystyle{IEEEtran}
\bibliography{myReferences}

\end{document}